\pdfoutput=1

\documentclass[11pt]{article}

\usepackage{ACL2023}

\usepackage{times}
\usepackage{latexsym}
\usepackage{arydshln}
\usepackage[T1]{fontenc}

\usepackage[utf8]{inputenc}

\usepackage{microtype}

\usepackage{inconsolata}
\usepackage{amsmath}
\usepackage{amsthm}
\usepackage{amsfonts}
\usepackage{multirow}
\usepackage{microtype}
\usepackage{graphicx}

\title{A Preliminary Empirical Study on Prompt-based Unsupervised \\Keyphrase Extraction}

\author{Mingyang Song, Yi Feng, Liping Jing\\
	Beijing Key Lab of Traffic Data Analysis and Mining \\
	Beijing Jiaotong University, Beijing, China \\
	{\tt mingyang.song@bjtu.edu.cn} \\
}

\begin{document}
\maketitle
\begin{abstract}
Pre-trained large language models can perform natural language processing downstream tasks by conditioning on human-designed prompts. However, a prompt-based approach often requires "prompt engineering" to design different prompts, primarily hand-crafted through laborious trial and error, requiring human intervention and expertise. It is a challenging problem when constructing a prompt-based keyphrase extraction method. Therefore, we investigate and study the effectiveness of different prompts on the keyphrase extraction task to verify the impact of the cherry-picked prompts on the performance of extracting keyphrases. Extensive experimental results on six benchmark keyphrase extraction datasets and different pre-trained large language models demonstrate that (1) designing complex prompts may not necessarily be more effective than designing simple prompts; (2) individual keyword changes in the designed prompts can affect the overall performance; (3) designing complex prompts achieve better performance than designing simple prompts when facing long documents.
\end{abstract}

\section{Introduction}
Keyphrase extraction aims at automatically extracting a set of phrases from the input document to summarize its core topics and primary information \cite{2014survey, song_survey}. Generally, keyphrase extraction models are trained on many document-keyphrase data pairs \cite{baseline, kiemp, csl, hypermatch}. These models demonstrate exceptional extractive capabilities for obtaining keyphrases from the given document, especially for Large Language Model (LLM) based keyphrase extraction systems. However, the quality of keyphrases extracted by prompt-based keyphrase extraction models is subject to the quality of the input prompts, whether in unsupervised or supervised settings. Designing proper prompts for keyphrase extraction models based on large pre-trained language models is challenging \cite{prompt1,song2,promptrank}.

\begin{figure}
	\centering
	\includegraphics[scale=0.42]{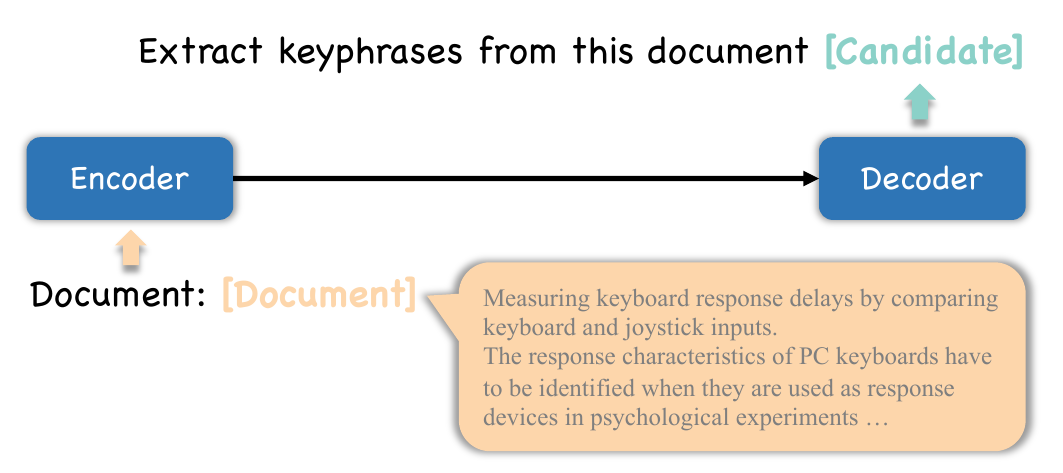}
	\caption{The illustration of a prompt-based keyphrase extraction model under an encoder-decoder architecture.}
	\label{instruction}
\end{figure}

In natural language processing, prompt-based learning is a new paradigm to replace fine-tuning large pre-trained language models on downstream tasks \cite{liupengfei}. Different from fine-tuning, prompt, the form of natural language, is more consistent with the pre-training task of models. Prompt-based learning has been widely used in many natural language processing tasks. In this paper, we analyze different prompts for unsupervised keyphrase extraction, leveraging the capability of large pre-trained language models with an encoder-decoder architecture.

As presented in Figure~\ref{instruction}, the general process of extracting keyphrases uses an encoder-decoder-based large pre-trained language model. It is necessary to design appropriate prompts to assist the model in outputting keyphrases for the input document, which means the design of prompts directly affects the performance of the prompt-based keyphrase extraction models. Typically, prompts for effectively extracting keyphrases are predominantly hand-crafted through laborious trial and error, requiring human intervention and expertise \cite{promptrank, song1, song2}. However, previous studies on keyphrase extraction has not systematically experimented with and analyzed whether complex or simple prompts might be more effective.

In this paper, we directly leverage a large pre-trained language model with an encoder-decoder architecture (i.e., T5 \cite{t5}) to measure the similarity without fine-tuning. Specifically, after extracting keyphrase candidates from the original document, we feed the input document into the encoder and calculate the probability of generating the candidate with a designed prompt by the decoder. The higher the probability, the more important the candidate. Experimental results on six benchmark keyphrase extraction datasets and different models demonstrate that (1) designing a complex prompt may not necessarily be more effective than designing a simple prompt; (2) individual keyword changes in a designed prompt can affect the overall performance; (3) designing a complex prompt achieve better performance than designing a simple prompt when facing long documents.

\section{Related Work}
Generally, unsupervised keyphrase extraction methods are divided into three categories: statistics-, graph-, and embedding-based models. Statistics-based models \cite{tfidf, yake} estimate the importance score of each candidate keyphrase by utilizing their statistical characteristics such as frequency, position, capitalization, and other features that capture the context information. The graph-based models \cite{textrank, topicrank, multipartiterank} are first proposed by TextRank \cite{textrank}, which treats each candidate keyphrase as a vertice, constructs edges according to the co-occurrence of candidates, and determines the weight of vertices through the PageRank algorithm. 

Embedding-based models \cite{keygames, sifrank, embedrank, icnlsp, setmatch, mderank} have achieved SOTA performance, benefiting from the recent development of the pre-trained language models, such as BERT \cite{bert} and RoBERTa \cite{roberta}. However, these algorithms perform poorly on long texts due to the length mismatch between the document and the candidate. \cite{mderank} solves the problem by replacing the embedding of the candidate with that of the masked document but fails to utilize the PLMs without fine-tuning fully. To address such issues,  \cite{promptrank} utilizes prompt-based learning for unsupervised keyphrase extraction.

In this paper, different from the existing models, we investigate the meaning of the design of prompts in the unsupervised keyphrase extraction task, leveraging the capability of pre-trained language models with an encoder-decoder architecture, such as T5 \cite{t5}.

\section{Methodology}
The main pipeline of prompting large language models for unsupervised keyphrase extraction is illustrated in Figure~\ref{instruction}. 
Following the recent work, we extract candidates from the document via heuristic rules. After obtaining candidates,  we first incorporate the document into a designed prompt as the input of the encoder and then calculate the probability of generating the candidate as the importance score with a designed prompt by the decoder. Finally, the importance score is used to rank and extract keyphrases. In analyzing the impact of different prompts in this paper, no additional parameter designs were introduced for fairness.

\begin{table*}[h]
	\scriptsize
	\centering
	\renewcommand\tabcolsep{11pt}
	\renewcommand\arraystretch{1.63}
	\begin{tabular}{ccccccc}
		\hline\hline
		\multirow{2}{*}{ \textbf{{Index}}} & \multirow{2}{*}{ \textbf{{Encoder}}} & \multirow{2}{*}{ \textbf{{Decoder}}} & \multirow{2}{*}{ \textbf{{Model}}} & \multicolumn{3}{c}{{\textbf{F1@K}}} \\  
		&    &   &   & \textbf{5} & \textbf{10} & \textbf{15}   \\ \hline
		\multirow{3}{*}{$p_1$}  & \multirow{3}{*}{"[Document]"}  & \multirow{3}{*}{"[Candidate]"} & \textsc{T5-base} & 11.46 & 16.62 & 18.68   \\ 
		& & &  \textsc{T5-3B}  & 11.36 & 16.52 & 18.83  \\ 
		& & &  \textsc{Flan-T5} & \textbf{11.56} & \textbf{16.80} & \textbf{19.19}  \\ \hline
		\multirow{3}{*}{$p_2$}  & \multirow{3}{*}{Article: "[Document]"}  & \multirow{3}{*}{This article mainly talks about "[Candidate]"} & \textsc{T5-base} & 15.76 & 21.74 & 23.35  \\ 
		& & &  \textsc{T5-3B} & \textbf{20.98} & \textbf{25.82} & \textbf{26.26}   \\ 
		& & &  \textsc{Flan-T5} & 18.46 & 23.37 & 24.20   \\ \hline
		\multirow{3}{*}{$p_3$}  & \multirow{3}{*}{Article: "[Document]"}  & \multirow{3}{*}{Keyphrases of this article are "[Candidate]"} & \textsc{T5-base} & 12.74 & 18.10 & 19.89  \\ 
		& & &  \textsc{T5-3B} & \textbf{14.46} & \textbf{19.72} & \textbf{21.57}  \\ 
		& & &  \textsc{Flan-T5} & 14.29 & 19.23 & 20.68   \\ 
		
		\hline\hline
	\end{tabular}
	
	\caption{The results of the same prompt with different keywords. "[Document]" is filled with the document, and "[Candidate]" is filled with the candidate. F1@K here is the average of six datasets.}
	\label{ori_prompt}
\end{table*}

\begin{table*}[h]
	\scriptsize
	\centering
	\renewcommand\tabcolsep{13pt}
	\renewcommand\arraystretch{1.66}
	\begin{tabular}{cccccc}
		\hline\hline
		\multirow{2}{*}{ \textbf{{Index}}} & \multirow{2}{*}{ \textbf{{Encoder}}} & \multirow{2}{*}{ \textbf{{Decoder}}} & \multicolumn{3}{c}{{\textbf{F1@K}}} \\  
		&   &   & \textbf{5} & \textbf{10} & \textbf{15}   \\ \hline
		$p_{1}$  & "[Document]"  & "[Candidate]"  & 11.46 & 16.62 & 18.68   \\ \cdashline{1-6}[1.5pt/2pt]
		$p_{1,1}$  &  \underline{Article}: "[Document]"  & "[Candidate]"  & 11.41 & 16.68 & 18.96   \\ 
		$p_{1,2}$  & "[Document]"  & \underline{Keyphrases}: "[Candidate]"  & 15.65 & 21.30 & \textbf{23.03}   \\ 
		$p_{1,3}$  & \underline{Article}: "[Document]"  & \underline{Keyphrases}: "[Candidate]"  & \textbf{16.42} & \textbf{21.39} & 22.88   \\ \hline
		$p_{2}$  & Article: "[Document]"  & This article mainly talks about "[Candidate]"  & 15.76 & 21.74 & 23.35   \\ \cdashline{1-6}[1.5pt/2pt]
		$p_{2,1}$  & \underline{Passage}: "[Document]"  & This \underline{passage} mainly talks about "[Candidate]"  & 15.55 & 21.30 & 23.04   \\ 
		$p_{2,2}$  & \underline{Book}: "[Document]"  & This \underline{book} mainly talks about "[Candidate]"  & 16.25 & \textbf{21.88} & 23.45   \\ 
		$p_{2,3}$  & \underline{Document}: "[Document]"  & This \underline{document} mainly talks about "[Candidate]" & 16.17 & 21.65 & 23.36   \\ 
		$p_{2,4}$  & \underline{Paper}: "[Document]"  & This \underline{paper} mainly talks about "[Candidate]"  & 16.06 & 21.66 & 23.28   \\ 
		$p_{2,5}$  & \underline{Content}: "[Document]"  & This \underline{content} mainly talks about "[Candidate]"  & 16.19 & 21.54 & 23.28   \\ 
		$p_{2,6}$  & \underline{Text}: "[Document]"  & This \underline{text} mainly talks about "[Candidate]"  & \textbf{16.34} & 21.87 & \textbf{23.49}   \\ \hline
		$p_{3}$  & Article: "[Document]"  & Keyphrases of this article are "[Candidate]"  & 12.74 & 18.10 & 19.89   \\ \cdashline{1-6}[1.5pt/2pt]
		$p_{3,1}$  & Article: "[Document]"  & \underline{Keywords} of this article are "[Candidate]"  & 13.31 & 18.39 & 19.90   \\ 
		$p_{3,2}$  & Article: "[Document]"  & \underline{The} keyphrases of this article are "[Candidate]"  & 13.61 & 18.82 & 20.52   \\ 
		$p_{3,3}$  & Article: "[Document]"  & \underline{Extract} keyphrases \underline{from} this article: "[Candidate]"  & \textbf{18.14} & \textbf{22.81} & \textbf{23.89}   \\ 
		\hline\hline
	\end{tabular}
	
	\caption{The performance of several prompts with different keywords. "[Document]" is filled with the document, and "[Candidate]" is filled with the candidate. F1@K here is the average of six datasets.}
	\label{prompts}
\end{table*}

\subsection{Candidate Extraction}
In this paper, we follow the previous studies and leverage the common practice \cite{hguke,mderank} to extract candidate keyphrases using the regular expression $<NN. *|JJ> * <NN.*>$ after tokenization and POS tagging.
\subsection{Importance Estimation}
Precisely, we fill the encoder template with the original input document and fill the decoder template with one candidate at a time. Then, we obtain the sequence probability $p(y_i|y_{<i})$ of the decoder template with the candidate based on pre-trained language models, such as T5 \cite{t5}. The length-normalized log-likelihood has been widely used due to its superior performance \cite{llm1}. Hence, we calculate the probability for one candidate as follows:
\begin{equation}
	\pi_c = - \frac{1}{l_c}\sum_{i=m}^{m+l_c-1}\log p(y_i | y_{<i}).
\end{equation}
where $l_c$ is the length of each candidate keyphrase. Here, we use $\pi_c$, whose value is positive, to evaluate the importance of candidates in ascending order. Then, select the top $K$ candidate keyphrases with the highest scores as the final set of keyphrases.

\begin{table*}[h]
	\scriptsize
	\centering
	\renewcommand\tabcolsep{9.9pt}
	\renewcommand\arraystretch{1.7}
	\begin{tabular}{ccccccccc}
		\hline\hline
		\multirow{2}{*}{ \textbf{F1@K}} & \multirow{2}{*}{ \textbf{{Model}}} & \multicolumn{6}{c}{\textsc{\textbf{Dataset}}} & \multirow{2}{*}{ \textbf{Avg.}} \\  
		&& {Inspec} & {SemEval2017} & SemEval2010 & DUC2001 & NUS & Krapivin &   \\ \hline
		\multirow{9}{*}{ F1@\textbf{{5}}}
		& \textsc{T5-base} ($p_{1,3}$) & 27.17 & 22.37 & 10.56 & 18.23 & 11.43 & 8.74 & 16.42 \\ 
		& \textsc{T5-base} ($p_{2,6}$) & \textbf{29.90} & \textbf{24.50} & \textbf{13.25} & \textbf{21.91} & 11.49 & \textbf{11.82} & \textbf{18.81} \\ 
		& \textsc{T5-base} ($p_{3,3}$) & 27.71 & 22.99 & 13.15 & 20.37 & \textbf{13.14} & 11.50 & 18.14 \\ \cdashline{2-9}[1.5pt/2pt]
		& \textsc{T5-3B}  ($p_{1,3}$)& 28.79 & 21.65 & 10.96 & 16.43 & 9.73 & 9.80 & 16.23 \\ 
		& \textsc{T5-3B}  ($p_{2,6}$)& \textbf{29.55} & \textbf{23.88} & \textbf{16.44} & \textbf{22.11} & \textbf{17.01} & \textbf{17.08} & \textbf{21.01} \\ 
		& \textsc{T5-3B}  ($p_{3,3}$)& 28.04 & 22.52 & 11.56 & 18.23 & 9.84 & 9.51 & 16.62 \\ \cdashline{2-9}[1.5pt/2pt]
		& \textsc{Flan-T5}  ($p_{1,3}$)& 28.28 & 21.77 & 10.76 & 12.05 & 9.50 & 9.35 & 15.29 \\
		& \textsc{Flan-T5}  ($p_{2,6}$)& \textbf{28.66} & \textbf{23.94} & \textbf{14.25} & \textbf{17.18} & \textbf{12.86} & \textbf{12.91} & \textbf{18.30} \\
		& \textsc{Flan-T5}  ($p_{3,3}$)& 28.54 & 22.17 & 11.46 & 12.60 & 9.56 & 9.43 & 15.63 \\\cline{2-9}

		\hline

		\multirow{9}{*}{ F1@\textbf{{10}}}
		& \textsc{T5-base} ($p_{1,3}$) & 33.65 & 31.19 & 15.56 & 22.79 & 14.31 & 10.83 & 21.39 \\ 
		& \textsc{T5-base} ($p_{2,6}$) & \textbf{35.45} & \textbf{33.82} & \textbf{18.03} & \textbf{25.64} & 15.01 & \textbf{13.42} & \textbf{23.56} \\ 
		& \textsc{T5-base} ($p_{3,3}$) & 33.75 & 31.82 & 17.63 & 23.87 & \textbf{16.45} & 13.37 & 22.81 \\ \cdashline{2-9}[1.5pt/2pt]
		& \textsc{T5-3B} ($p_{1,3}$) & \textbf{35.15} & 31.55 & 17.15 & 22.21 & 14.13 & 11.35 & 21.92 \\ 
		& \textsc{T5-3B} ($p_{2,6}$) & 34.58 & \textbf{33.72} & \textbf{19.39} & \textbf{26.86} & \textbf{20.26} & \textbf{17.37} & \textbf{25.36} \\ 
		& \textsc{T5-3B} ($p_{3,3}$) & 34.60 & 32.40 & 16.91 & 23.33 & 14.22 & 11.02 & 22.08 \\ \cdashline{2-9}[1.5pt/2pt]
		& \textsc{Flan-T5} ($p_{1,3}$) & 34.52 & 30.91 & 17.15 & 16.73 & 13.48 & 10.94 & 20.62 \\
		& \textsc{Flan-T5} ($p_{2,6}$) & 34.02 & \textbf{32.95} & \textbf{18.91} & \textbf{21.20} & \textbf{16.63} & \textbf{14.17} & \textbf{22.98} \\
		& \textsc{Flan-T5} ($p_{3,3}$) & \textbf{34.85} & 31.02 & 17.07 & 17.96 & 13.56 & 10.83 & 20.88 \\ \cline{2-9}

		\hline

		\multirow{9}{*}{ F1@\textbf{{15}}}
		& \textsc{T5-base} ($p_{1,3}$) & 33.95 & 34.85 & 17.69 & 23.89 & 15.61 & 11.30 & 22.88 \\ 
		& \textsc{T5-base} ($p_{2,6}$) & \textbf{34.82} & \textbf{37.33} & \textbf{18.82} & \textbf{26.15} & 16.32 & \textbf{13.37} & \textbf{24.47} \\ 
		& \textsc{T5-base} ($p_{3,3}$) & 33.79 & 35.54 & \textbf{18.82} & 24.91 & \textbf{17.31} & {13.00} & {23.89} \\ \cdashline{2-9}[1.5pt/2pt]
		& \textsc{T5-3B} ($p_{1,3}$) & \textbf{35.38} & 35.10 & 18.96 & 23.35 & 16.35 & 11.74 & 23.48 \\ 
		& \textsc{T5-3B} ($p_{2,6}$) & 34.51 & \textbf{36.54} & \textbf{20.42} & \textbf{27.54} & \textbf{20.76} & \textbf{15.93} & \textbf{25.95} \\ 
		& \textsc{T5-3B} ($p_{3,3}$) & 35.08 & {35.88} & 18.96 & 24.71 & 15.39 & 11.68 & 23.62 \\ \cdashline{2-9}[1.5pt/2pt]
		& \textsc{Flan-T5} ($p_{1,3}$) & \textbf{35.02} & 34.69 & 19.42 & 19.35 & 15.14 & 11.21 & 22.47 \\
		& \textsc{Flan-T5} ($p_{2,6}$) & 33.93 & \textbf{36.10} & \textbf{20.09} & \textbf{22.84} & \textbf{17.67} & \textbf{13.14} & \textbf{23.96} \\
		& \textsc{Flan-T5} ($p_{3,3}$) & 34.90 & 34.61 & 19.62 & 19.72 & 15.36 & 10.56 & 22.46 \\ \cline{2-9}

		\hline\hline
	\end{tabular}
	
	\caption{The performance of keyphrase extraction on six datasets. The best results are highlighted in bold.}
	\label{present}
\end{table*}

\section{Experiment}
We present the used datasets and evaluation metrics, the implementation details, and the results.

\subsection{Datasets}
In this paper, we conduct experiments on six widely used keyphrase extraction benchmark datasets, such as Inspec \cite{Inspec}, DUC2001 \cite{duc2001_singlerank}, SemEval2010 \cite{semeval2010}, SemEval2017 \cite{semeval2017}, Nus \cite{Nus}, and Krapivin \cite{Krapivin}.
\subsection{Evaluation Metrics}
Following the previous researches \cite{setmatch, hguke, song-etal-2023-mitigating, song-etal-2023-hyperrank, promptrank}, we adopt F1 on the top 5, 10, and 15 ranked candidates to evaluate the results in this paper. When calculating F1 score, duplicate candidate keyphrases are removed, and stemming is applied.
\subsection{Implementation Details}
We adopt the pre-trained language model T5 \cite{t5} as the backbone, initialized from their pre-trained weights. Among them, there are two versions used in this paper, such as "T5-base" and "T5-3B". Furthermore, we also use the pre-trained language model Flan-T5-base \cite{flan-t5} as the backbone to conduct experiments.
Similar to the recent work, to match the settings of BERT \cite{bert}, the maximum length for the inputs of the encoder is set to 512. In addition, we utilize the code from \citet{promptrank} to complete the experiments in this paper. The difference is that we do not introduce any adjustable parameters. For more details, please refer to \citet{promptrank}.
\subsection{Results}

As mentioned before, we mainly focus on investigating and studying the effectiveness of different prompts on the keyphrase extraction task to verify the impact of the cherry-picked prompts on the performance of extracting keyphrases in this paper. Therefore, we design three types of prompts (ranging from simple to complex) suitable for extracting keywords. Then, we conduct experiments on different large pre-trained language models, further replace keywords in prompts, and analyze the necessity of cherry-picked prompts. All results are displayed in Table~\ref{ori_prompt}, Table~\ref{prompts}, and Table~\ref{present}. Next, we analyze the experimental results in detail.

From the results in Table~\ref{ori_prompt}, it can be seen that when the prompt ($p_1$) is not used at all, the performance of T5-base and T5-3b are both poor, and even the performance of T5-3b is not as good as T5-base, while Flan-T5 achieved the best effect. After using more detailed prompts ($p_2$ and $p_3$), it can be found that the results of T5-3B and Flan-T5 are significantly better than those of the T5-base. Furthermore, it can be seen from Table~\ref{ori_prompt} that using $p_2$ as a prompt can achieve better performance than using $p_3$ as a prompt, whether using T5-base, T5-3B, or Flan-T5.

Many existing approaches attempt to construct various prompts, such as modifying different keywords in the prompts, to achieve better performance. Hence, we also analyzed the impact of different keywords in the modification prompt on the results. Taking inspiration from existing methods \cite{promptrank, song1, song2}, we modified the keywords in the three prompts used in this paper and verified their performance. The results are shown in Table~\ref{prompts}. From the results, we can find that the designed prompts $(p_{1,3}, p_{2,6}, p_{3,3})$ obtain the best results, respectively. However, we found that changing different keywords has little effect on the results in most cases, indirectly indicating the effectiveness of constructing refined prompts but requiring a lot of experimentation.

The results in Table~\ref{present} show that T5-3B performs significantly better than T5-base and Flan-T5 on long document datasets, such as the SemEval2010 dataset. Meanwhile, the results of $p_{2,6}$ are considerably better than those of $p_{1,3}$ and $p_{3,3}$, which indicates the necessity of designing a complex prompt. The difference in the results is insignificant with different prompts, so a refined prompt design is not a reasonable strategy based on existing results. On the contrary, automatic prompt generation or search should be more convenient and efficient.

\section{Conclusion}

In this paper, we investigate the effectiveness of different prompts to verify the impact of the cherry-picked prompts on the performance of extracting keyphrases. Extensive experimental results on six benchmark keyphrase extraction datasets and different pre-trained large language models demonstrate that (1) designing complex prompts may not necessarily be more effective than designing simple prompts in most cases; (2) individual keyword changes in prompts affect the overall performance; (3) designing complex prompts achieve better performance than designing simple prompts when facing long documents. Future research may be possible to better extend similar ideas from phrase-level to sentence-level information extraction (i.e., the extractive summarization task \cite{hypersiamesenet, hisum}) in the future. In addition, it might be possible to construct a new long-context benchmark, such as the needle-in-a-haystack\footnote{\url{https://github.com/gkamradt/LLMTest_NeedleInAHaystack}} or the Counting-Stars \cite{song2024countingstars}, through the keyphrase extraction task.

\bibliography{anthology}
\bibliographystyle{acl_natbib}
\end{document}